\documentclass{ws-jmrr}
\usepackage[sort,compress,super]{cite}
\usepackage{gensymb}
\usepackage{amsmath,mathtools}
\usepackage{xcolor}

\begin{document}

\catchline{0}{0}{2013}{}{}

\markboth{Behnam Moradkhani}{ExoNav II: Robotic Tool with Follow-the-Leader Motion Capability for \textcolor{black}{Lateral and} Ventral Spinal Cord Stimulation (SCS)}

\title{ExoNav II: Design of a Robotic Tool with Follow-the-Leader Motion Capability for \textcolor{black}{Lateral and} Ventral Spinal Cord Stimulation (SCS)}

\author{Behnam Moradkhani$^a$ , Pejman Kheradmand$^a$, Harshith Jella$^a$, Joseph Klein$^a$,\\ Ajmal Zemmar$^{\textcolor{black}{b,*}}$ and Yash Chitalia$^{a,\textcolor{black}{b},}$\textcolor{black}{\footnote{\textcolor{black}{Equal senior author contribution.}}}}

\address{$^a$Healthcare Robotics and Telesurgery (HeaRT) Laboratory, J. B. Speed School of Engineering, University of Louisville, Louisville, Kentucky, USA\\
E-mail: b0mora01@louisville.edu}

\address{$^b$Department of Neurological Surgery, University of Louisville School of Medicine, Louisville, Kentucky, USA.}

\maketitle

\begin{abstract}
Spinal cord stimulation (SCS) electrodes are \textcolor{black}{traditionally} placed in the \textcolor{black}{dorsal} epidural space \textcolor{black}{to stimulate the dorsal column fibers for pain therapy. Recently, SCS has gained attention in restoring gait. However, the motor fibers triggering locomotion are located in the ventral and lateral spinal cord. Currently, SCS electrodes are steered manually, making it difficult to navigate them to the lateral and ventral motor fibers in the spinal cord.} In this work, we propose a helically micro-machined continuum robot that can bend in a helical shape when subjected to actuation tendon forces. Using a stiff outer tube and adding translational and rotational degrees of freedom, this helical continuum robot can perform follow-the-leader (FTL) motion. We propose a kinematic model to relate tendon stroke and geometric parameters of the robot's helical shape to its acquired trajectory and end-effector position. We evaluate the proposed kinematic model and the robot's \textcolor{black}{FTL} motion capability \textcolor{black}{experimentally}. \textcolor{black}{The stroke-based method, which links tendon stroke values to the robot's shape, showed inaccuracies with a 19.84 mm deviation and an RMSE of 14.42 mm for 63.6 mm of robot's length bending. The position-based method, using kinematic equations to map joint space to task space, performed better with a 10.54 mm deviation and an RMSE of 8.04 mm. Follow-the-leader experiments showed deviations of 11.24 mm and 7.32 mm, with RMSE values of 8.67 mm and 5.18 mm for the stroke-based and position-based methods, respectively.} \textcolor{black}{Furthermore, end-effector trajectories in two FTL motion trials are compared to confirm the robot's repeatable behavior.} Finally, we demonstrate the robot's operation on a 3D-printed spinal cord phantom model.
\end{abstract}


\begin{multicols}{2}
\section{Introduction}
Epidural spinal cord stimulation (SCS) is a therapeutic approach primarily used for patients with chronic pain who have not responded to other treatments [\citen{knotkova2021neuromodulation}, \citen{VINUELAPRIETO2021100301}], demonstrating significant efficacy in reducing pain. Beyond pain \textcolor{black}{therapy}, SCS has shown promising results in restoring voluntary motor functions in spinal cord injury (SCI) patients [\citen{jcm13041090}]. Additionally, SCS \textcolor{black}{can activate proprioceptive function and} autonomic function recovery, \textcolor{black}{including} bladder control, blood pressure regulation, and respiratory function [\citen{jcm13041090}]. However, traditional SCS primarily targets the dorsal spinal cord for accessibility, limiting its efficacy in stimulating motor neurons \textcolor{black}{(located at lateral and ventral epidural space proximity) that activate movement and proprioceptive fibers. Maneuvering electrodes around the epidural space is challenging and involves significant risk with the current standard of manual steering} [\citen{VStimulation}].
 \textcolor{black}{Various methods} have been \textcolor{black}{proposed} to enhance precision and accessibility for dorsal SCS electrode placement, mainly for pain \textcolor{black}{therapy}. To improve electrode placement, MacDonald and Fisher introduced a suture retrieval snare for midline or paramedian positioning \textcolor{black}{[\citen{macdonald2011technique}]}, and Loge et. al. used natural fibrous deposits to facilitate electrode replacement \textcolor{black}{[\citen{loge2002epidural}]}. However, these approaches rely on fixed shapes for navigation and lack steering capabilities. Torlakcik et. al. proposed a magnetically guided catheter for real-time navigation control using an external magnet \textcolor{black}{[\citen{torlakcik2021magnetically}]}. In our previous work, we introduced ExoNav, an innovative steering robot designed for precise SCS lead placement while minimizing risks of collisions with delicate anatomical structures \textcolor{black}{[\citen{jella2024exonav}]}.
\textcolor{black}{However, robotic solutions selectively exploring navigation to lateral and ventral fiber tracts have yet not been investigated.} In this regard, proposed robotic solutions must address several critical hurdles. First, the robot’s design must meet strict size constraints to fit within the limited epidural space. Second, the robot’s shape and maneuverability must enable it to reach the ventral \textcolor{black}{and lateral spinal cord} from a dorsal entry point, \textcolor{black}{thus} safely navigating around the spinal cord. Finally, the robotic system must be capable of performing a precise “follow-the-leader” (FTL) motion, which is essential for controlled trajectory-based navigation in this sensitive anatomical context. These challenges may explain the limited efforts in \textcolor{black}{robotic systems targeting ventral and lateral spinal cord}.\\
\textcolor{black}{First introduced around four decades ago \textcolor{black}{[\citen{choset1999follow}]}, FTL motion has become a prominent research challenge in robotics, with increased focus on it over the last decade \textcolor{black}{[\citen{culmone2021follow}]}. This technique involves a type of movement where the centerline of the device follows the tip, traveling along a predetermined curved trajectory, ensuring that the surrounding environment remains undisturbed. Various robotic \textcolor{black}{(and non-robotic [\citen{henselmans2020memoflex}])} designs capable of FTL motion \textcolor{black}{have been} proposed for diverse applications, including mobile modular robots \textcolor{black}{[\citen{SnakeLikeRobotFTL, TappeOptimization, komura2015development, chen2014modular}]}, shape memory alloy (SMA) robots \textcolor{black}{[\citen{palmer2014real}]}, soft inflatable robots \textcolor{black}{[\citen{hawkes2017soft, coad2020retraction}]}\textcolor{black}{, and continuum robots} \textcolor{black}{[\citen{neumann2016considerations, jeong2020design, EfficientFTLMotion, FTLMedicalDevices, SLAM-basedGirerd}]}. Such motion is particularly beneficial in surgical settings when navigating sensitive anatomies, like the spinal cord. Continuum surgical robots facilitate minimally invasive access to complex regions, enhancing precision in procedures like laparoscopy and endoscopy, while reducing recovery times and easing surgical workflows [\citen{da2020challenges, omisore2020review}].} \textcolor{black}{As a result,} several studies have explored the implementation of FTL motion in continuum robots, aiming to enhance precision and reduce risks in delicate surgeries.\\
 Tendon-driven \textcolor{black}{continuum robot} designs are particularly favored in medical robotics for their high controllability and fast actuation \textcolor{black}{[\citen{hu2018steerable}]}. Among these, tendon-driven robots capable of FTL motion fall into two main categories: multi-segment robots with extensible disks \textcolor{black}{[\citen{neumann2016considerations}]} and micromachined tube robots \textcolor{black}{[\citen{jeong2020design}]}. While the former faces challenges in miniaturization and often relies on permanent ring magnets (lacking MR-compatibility), micromachined tubes (often made from MR-compatible nitinol \textcolor{black}{[\citen{melzer2004Nitinol}]}) offer significant miniaturization and high curvature, making them ideal for precise SCS navigation in the epidural space.\\
In this study, we propose \textbf{\textit{the very first robotic steering method \textcolor{black}{to directly navigate to ventral and lateral SCS fibers targeting} motor function control in SCI patients}}. The design of the proposed robot is intentionally compact to \textcolor{black}{suit the confinement of the} epidural space and adaptable to the specific needs of this application. This approach is an updated version of the ExoNav robot, initially designed for dorsal SCS and pain management \textcolor{black}{[\citen{jella2024exonav}]}. In this new ExoNav robot, we engineer the robotic tool to adopt helical shape when actuated, which is ideal for \textcolor{black}{and naturally capable of} performing FTL motion through simultaneous rotation and translation \textcolor{black}{[\citen{henselmans2019mechanical}]}, ensuring safe and precise navigation to the ventral side of the spinal cord. We analyze the geometry of the proposed robot and develop a kinematic model. Ultimately, we evaluate the derived kinematic equations experimentally\textcolor{black}{, FTL motion capabilities, operational repeatability, and the robot's function in a phantom model environment}.
This manuscript is organized as follows: In Section \ref{sec2:Design}, we present a comprehensive design description of the ExoNav robot. In Section \ref{sec3:kinematics}, we elaborate upon the geometry of a cylindrical helix and relate it to the proposed design to develop a forward kinematic model. In Section \ref{sec4:Experiments}, we test our model via multiple experimental scenarios and analyze its accuracy in estimating the robot's behavior, \textcolor{black}{and its repeatability in trajectory-following navigation,} along with a phantom model demonstration. Finally, Section \ref{sec5:conclusion} concludes this work with a discussion of future directions.
\section{Design} \label{sec2:Design}
\begin{figure*}[t]
\centerline{\includegraphics[width=6.1in,keepaspectratio]{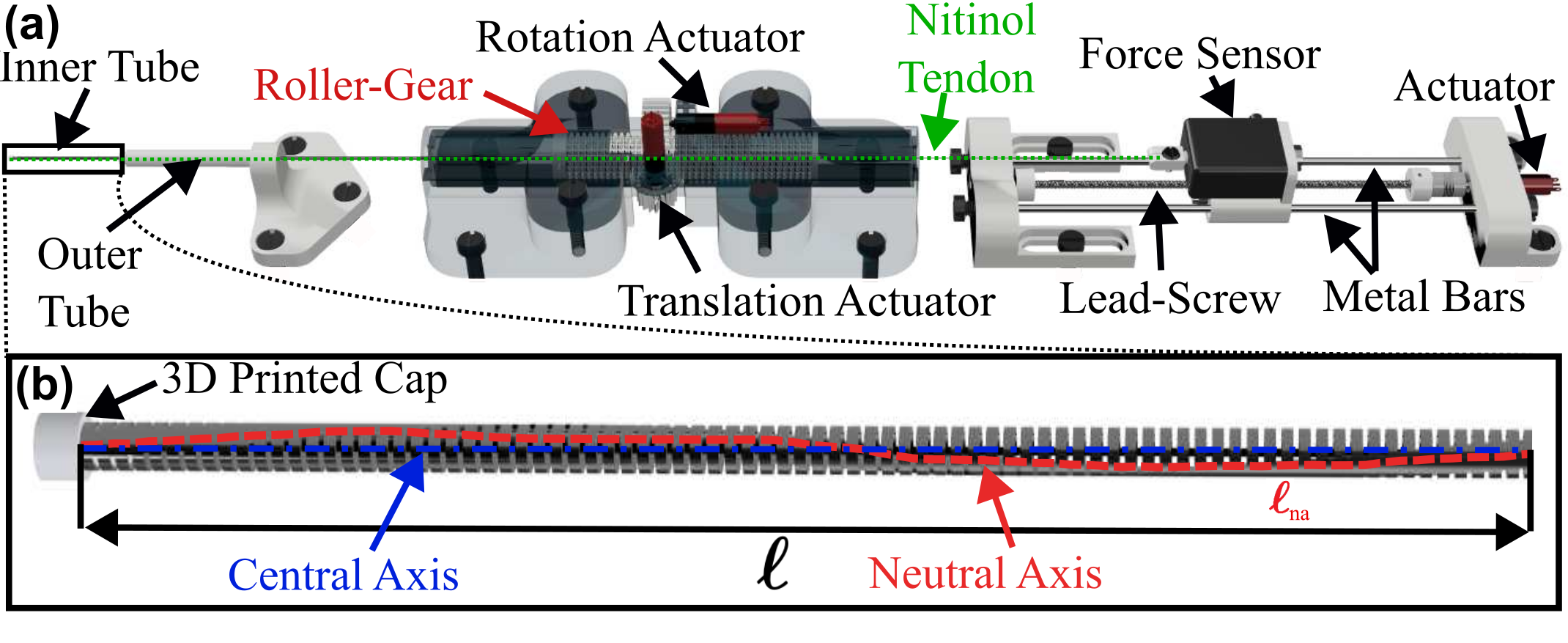}}
\caption{(a) ExoNav robotic system design with inner and outer tubes, the roller gear and linear tendon pulling mechanisms at the back-end (left-to-right), (b) close-up view of the inner tube with helical cut patterns and a 3D-printed cap, utilized for anchoring the actuation tendon, installed at the tip of the robotic tool.}
\label{fig:Design}
\end{figure*}
\textcolor{black}{ExoNav} consists of two parts \textcolor{black}{(see Fig. \ref{fig:Design})}: 1) a robotic tool with a helical-patterned nitinol inner tube telescoped within a 3D-printed outer tube, 2) a back-end actuation unit to translate \textcolor{black}{and} rotate \textcolor{black}{the inner tube with respect to the fixed outer tube}, and tension the \textcolor{black}{actuation tendon, bending the inner tube to reach the desired helical shape}.
\subsection{Robotic Tool with Helical Cut Pattern}
Two tubes form the ExoNav robot: A nickel-titanium (nitinol) inner tube, and a 3D-printed outer tube. The inner tube is a machined nitinol tube (with inner and outer radii of $R_{in}=0.851$~mm and $R_{out}=0.953$~mm\textcolor{black}{, as provided by the manufacturer}) with a specific quasi-helical micromachining pattern that allows the robotic tool to deform \textcolor{black}{into} a helical shape. The cut pattern machined along the tube \textcolor{black}{over a length of $l=64$~mm}, is a \textcolor{black}{set of} rectangular shapes with edge dimensions of $w=0.5$~mm and $h=3.892$~mm, which are parallel and perpendicular to the inner tube's central axis respectively (See Fig. \ref{fig:NeutralAxis}(a)-(b)). Each rectangle is axially distanced from the next rectangular cut by a bridge of tube material of length $d=0.3$~mm (See Fig. \ref{fig:NeutralAxis}(b)) and is offset from the next one by distance $a=0.075$~mm along the circumference of the tube (See Fig. \ref{fig:NeutralAxis}(a)). These notches are \textcolor{black}{precisely} made along the distal length of a nitinol tube using a femtosecond laser (Optec WS-Flex, Institute for Electronics and Nanotechnology, Georgia Institute of Technology, Atlanta) and the overall pattern resembles a helix drawn along the outer surface of the tube. A nitinol tendon with radius of $r_t=0.115$~mm is \textcolor{black}{then} routed through \textcolor{black}{the inner} tube and is \textcolor{black}{constrained} at the distal end to a 3D-printed cap placed at \textcolor{black}{the tip} of the \textcolor{black}{inner} tube (See Fig. \ref{fig:Design}). \textcolor{black}{This 3D-printed cap is a component with a small hole through which the tendon is passed and knotted at the tip, so it gets anchored to the cap while being tensioned. Note that the constrained tendon knot is free to rotate if needed and effects of friction at the distal attachment point are insignificant.} Actuating this tendon will generate forces and moments at the tip of the \textcolor{black}{inner tube} thereby deforming it in a helical configuration. This will be further analyzed in Section \ref{sec3:kinematics}. The \textcolor{black}{inner} tube with its tendon installed is then placed inside a 3D-printed outer tube. This outer tube prevents deformation of \textcolor{black}{the inner tube's} proximal section. Therefore, the length of the inner tube that is exposed out of the outer tube can deform freely and the length that is within the outer tube will remain \textcolor{black}{straight and} undeformed.
\vspace{-10pt}
\subsection{Back-end Actuation Unit}
The back-end actuation unit is made of two components\textcolor{black}{: 1) a linear actuator with a lead-screw coupled to a DC motor (RE 8 Ø8 mm, Precious Metal Brushes, 0.5 Watt, with terminals, Maxon international ltd., Switzerland) that can move a bracket along two installed steel rods, and 2) a dual degree-of-freedom (DoF)} 3D-printed (ProJet MJP 2500, 3D Systems Inc.) roller gear with circumferential and longitudinal involute gear profiles that allow for simultaneous rotation and translation \textcolor{black}{[\citen{morimoto2017design}]} (See Fig. \ref{fig:Design}(a)). The \textcolor{black}{inner tube's proximal end (without machined pattern)} is attached to the tip of the roller gear. \textcolor{black}{Therefore, the inner tube can be translated and rotated with respect to the fixed outer tube, by the motors installed on the roller gear mechanism, which are equipped with incremental encoders to measure the displacement and rotation applied}. The tendon controlling the inner tube of the ExoNav robot is routed centrally through the roller gear mechanism and fixed to the moving bracket on the linear actuator \textcolor{black}{(axially aligned in position at the back of the roller gear mechanism)}, which is equipped with a tendon tension sensor (MDB-5 force sensor, Transducer Techniques). A Texas Instrument LAUNCHXL-F28379D LaunchPad is utilized for motor control and to gather force and motor encoder data in the MATLAB/Simulink environment.
\section{Kinematics}\label{sec3:kinematics}
In this section, we present a kinematic model based on the geometric shape of a cylindrical helix. This model defines the relationship \textcolor{black}{among} actuation tendon stroke $\Delta l_{t}$ (actuation space), the helical shape of the robot's inner tube and the cylinder that the helix wraps around (joint space), and the position vectors to arbitrary points along the inner tube's length, expressed as a function of the arc length along the \textcolor{black}{inner tube}'s body (task space).
\subsection{Neutral Axis of the Inner Tube}
\begin{figurehere}
\begin{center}
\centerline{\includegraphics[width=3in,keepaspectratio]{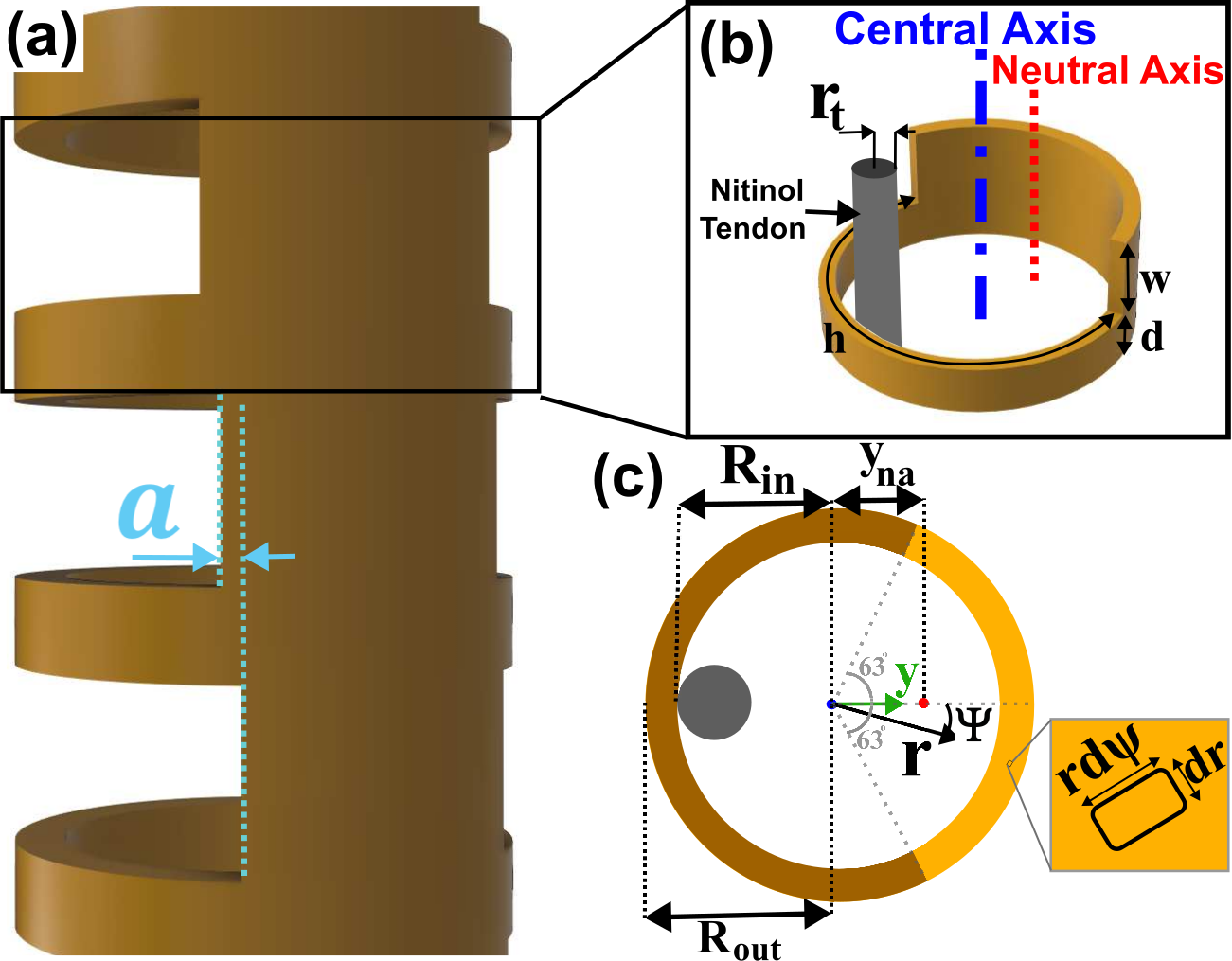}}
\caption{(a) Section of the inner tube showing the rectangular cut patterns and the offset $a$ between consecutive cuts, (b) Isometric view of a separated segment of the inner tube with neutral axis, central axis, and the nitinol tendon, (c) Top view of the separated segment of the inner tube.}
\label{fig:NeutralAxis}
\end{center}
\end{figurehere}
 Due to the quasi-helical pattern micromachined with rectangular notches within the inner tube, the neutral axis of the \textcolor{black}{unactuated} inner tube also forms a helix (See Fig. \ref{fig:Design}(b)). To determine the location of the neutral axis inside the inner tube, we focus on a single segment of the inner tube that has a rectangular notch, and a solid segment that connects consecutive notches. Using the polar coordinates shown in Fig. \ref{fig:NeutralAxis}(c) with radial and angular variables $r$ and $\psi$ respectively, the distance of the partial neutral axis for the represented segment from the central axis of the tube is denoted by $y_{na}$ (see Fig. \ref{fig:NeutralAxis}(c)).
\begin{equation} \label{NA_general_formula}
     y_{na}= \frac{\int_{A} y \,dA}{\int_{A} \,dA}
 \end{equation}
  The parameter $y$ is the horizontal axis which can be written as $r\cos(\psi)$. Additionally, $A$ is the area of the material on the plane, and \textcolor{black}{focusing only on the notched portion of the inner tube's separated segment}, it is the arc-shaped surface that extends from $\psi=-63\degree$ to $\psi=+63\degree$ in angular direction, and from $r=R_{in}=0.851$~mm to $r=R_{out}=0.953$~mm in radial direction. The differential segment of area is considered a rectangle with edges of $rd\psi$ and $dr$ (See Fig. \ref{fig:NeutralAxis}(c)). So, Eq. (\ref{NA_general_formula}) can be rewritten as:
\begin{equation} \label{NA_formula}
     \textcolor{black}{y_{na}^{notch}}=\frac{\int_{R_{in}}^{R_{out}} \int_{-63\degree}^{+63\degree} r^2 \cos(\psi) \,d\psi \,dr}{\int_{R_{in}}^{R_{out}} \int_{-63\degree}^{+63\degree} r \,d\psi \,dr}
 \end{equation}
 Solution of Eq. (\ref{NA_formula}) results in the neutral axis location of the notch at \textcolor{black}{$y_{na}^{notch}=0.7318$~mm} from the central axis. \textcolor{black}{The overall neutral axis distance from the central axis ($y_{na}$) is then calculated by superposing the distances obtained for the notched and unnotched portions ($y_{na}^{notch}$ and $y_{na}^{bridge}=0$, respectively)}.
 \begin{equation}
     \textcolor{black}{y_{na} = \frac{wy_{na}^{notch}+dy_{na}^{bridge}}{w+d}}
 \end{equation}
 \textcolor{black}{So, the overall neutral axis location} is $y_{na}=0.4574$~mm. Noting that the neutral axis is a helix, the length of the neutral axis $l_{na}$ can be obtained as:
\begin{equation} \label{l_na}
    l_{na} = \sqrt{l^2 + (2\pi y_{na})^2}
\end{equation}
In developing this model, we consider the case where the inner tube makes one entire helical rotation around the spinal cord. However, in general, the number of turns the \textcolor{black}{inner tube} can make around any arbitrary cylinder (proxy for the spinal cord) will be equal to the number of helical turns micro-machined within the nitinol tube.
\subsection{Inner Tube Shape}
The shape of the ExoNav robot's inner tube is a function of the tendon stroke $\Delta l_t$. We define an imaginary cylinder which determines the overall shape of the \textcolor{black}{inner tube} (See Fig. \ref{fig:Kinematics}(a)). We assume that when deformed, the neutral axis of the inner tube wraps around this imaginary cylinder (See Fig. \ref{fig:Kinematics}). 
\begin{equation} \label{l_na_R_H}
    l_{na} = \sqrt{H^2 + (2\pi R)^2)}
\end{equation}
 As we observed, the actuation tendon is taking the shortest possible path inside the inner tube (attaches to the notched side of the tube) and stays there while the inner tube is actuated and the tube deforms into a helical shape (See the tendon location near the notched side inside the inner tube in Fig. \ref{fig:NeutralAxis}(b) and (c)). Therefore, the tendon length $l_t$ is obtained as a function of the imaginary cylinder dimensions ($R$ and $H$, as shown in Fig. \ref{fig:Kinematics}(a)) and the fixed distance between the actuation tendon and the neutral axis which is $d_{t-na}=y_{na}+r_{in}-r_{t}$.
\begin{equation} \label{l_t_R_H}
    l_t = \sqrt{H^2 + (2\pi (R-d_{t-na}))^2}
\end{equation}
Since $l_{na}$ is fixed, Eq. (\ref{l_na_R_H}) shows there are specific combinations of $R$ and $H$ parameters that can be achieved with a specific \textcolor{black}{inner tube's notch pattern} design. Using Eqs. (\ref{l_t_R_H}) and (\ref{l_na_R_H}), Radius of the imaginary cylinder is obtained.
\begin{equation} \label{R-l_t relation}
    R = \frac{l_{na}^2-l_{t}^2}{8\pi^2 d_{t-na}}+\frac{d_{t-na}}{2}
\end{equation}
Substituting the obtained parameter $R$ into Eq. (\ref{l_t_R_H}), the height of the imaginary cylinder $H$ is obtained. Note that the length of the tendon $l_t$ can suitably be related to $\Delta l_t$ and the original length of the tendon $l_{t0}$ routed along the micromachined portion of the inner tube.
\begin{equation} \label{l_t to Delta l_t relation}
    l_t = l_{t0} - \Delta l_t + l_e
\end{equation}
The parameter $l_e$ is the elongation that the tendon experiences and it depends on the tendon tension $T$ (measured experimentally by the tendon tension sensor), total length of the tendon in use $l_{t-total}=475$~mm (length of the green dotted line in Fig. \ref{fig:Design}(a)), area of the tendon's cross-section $A_t=1.135\times 10^{-6}$~m$^2$, and the elastic modulus of the tendon ($E_t=53.97$~GPa according to \textcolor{black}{[\citen{jeong2020design}]}).
\begin{equation}
    l_e = \frac{T l_{t-total}}{A_t E_t}
\end{equation}
Therefore, we derive the dimensions of the imaginary cylinder purely based on the applied tendon stroke values.
\subsection{Actuation and Deflection Angles}
Fig. \ref{fig:Kinematics}(a) illustrates that the inner tube's helix deflects from its straight configuration as it exits the outer tube by a deflection angle $\phi$. Since the inner tube must stay tangent to the outer tube path at the point where the inner tube extends out, this angle gets introduced upon tendon actuation. Noting the changes made in the imaginary cylinder dimensions, the deflection angle is obtained as follows:
\begin{equation} \label{phi-R-H}
    \phi = \arctan (\frac{2\pi (R-y_{na})}{H})
\end{equation}
Additionally, there is another angle involved in determining the shape of the \textcolor{black}{inner tube} which is called the actuation angle $\theta$. This angle is directly controlled by the roller gear's rotational DoF. It simply determines how much the inner tube has been rotated with respect to a fixed reference. In other words, it determines the direction in which the inner tube's notches are facing as they appear at the tip of the outer tube. It also determines the direction in which the inner tube will deflect. The actuation angle $\theta$, along with the deflection angle $\phi$, is used to define the coordinate systems that obtain the shape and location of the robotic tool (See Fig. \ref{fig:Kinematics}(b)).
\subsection{Position Vectors in 3D Space}
\begin{figurehere}
\begin{center}
\centerline{\includegraphics[width=3in,keepaspectratio]{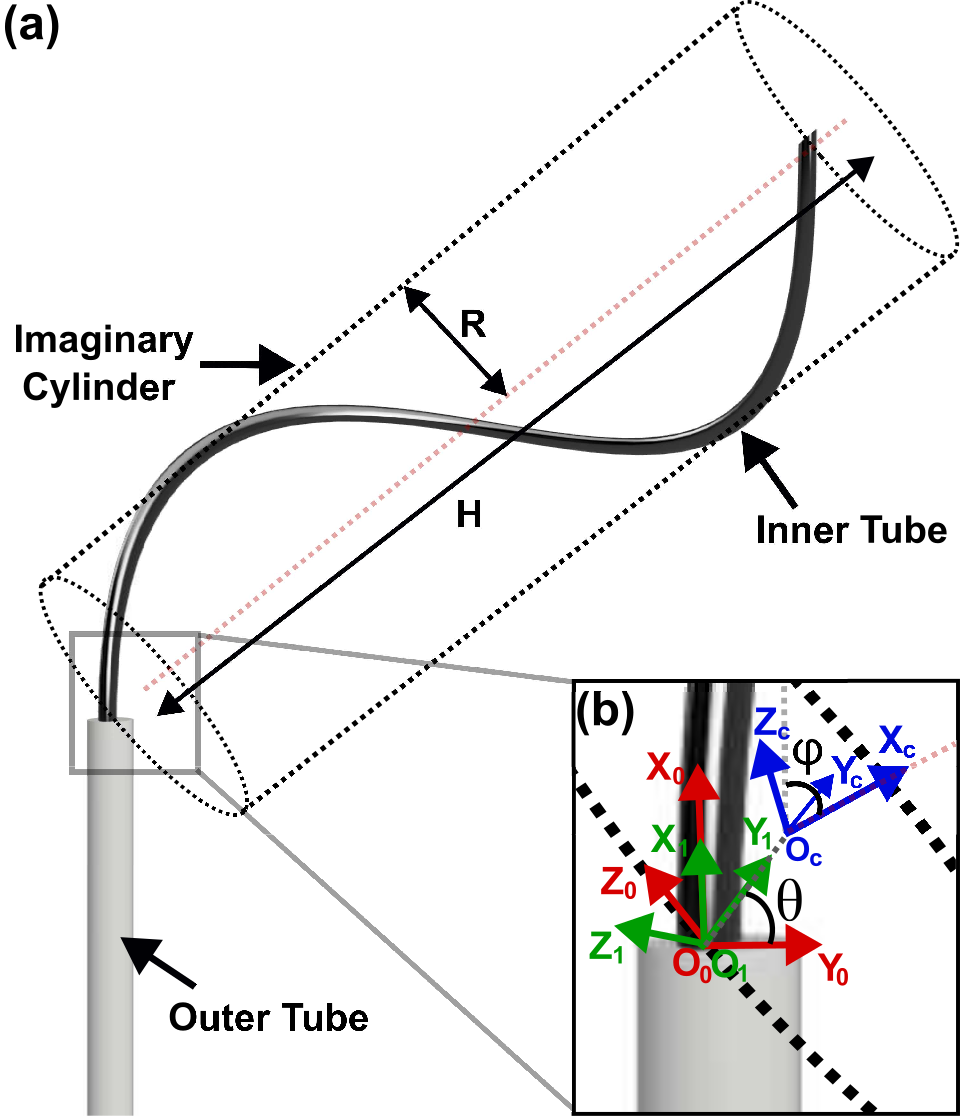}}
\caption{(a) Actuated inner tube bends in a helical shape and forms an imaginary cylinder, (b) Coordinate frames defined at the tip of the outer tube and at the center point of the bottom circle of the imaginary cylinder, with respective angles between frames highlighted.}
\label{fig:Kinematics}
\end{center}
\end{figurehere}
\vspace{-25pt}
As a fixed global coordinate system, $O_0$ is defined at the tip of the outer tube and on the inner tube's neutral axis with $X_0$ pointed out and along the outer tube and with other axes directions chosen arbitrarily. The local coordinate system $O_1$ is defined at the same origin location as $O_0$ and utilized to define the actuation angle, $\theta$. Therefore, by definition, the axis $Y_1$ is pointed towards the notches at any arbitrary position along the inner tube's arc length. Therefore, at the point where the inner tube exits the outer tube, $Y_1$ makes an angle equal to $\theta$ with $Y_0$ about $X_0$ (See Fig. \ref{fig:Kinematics}(b)). With $Y_1$ facing the same direction as the notches on the inner tube, the bending of the \textcolor{black}{inner tube} will be in a way that the center of the base of our imaginary cylinder is at a distance $R$ from $O_1$, along $Y_1$. This location is where the coordinate system $O_c$ is defined (See Fig. \ref{fig:Kinematics}(b)). In this coordinate system, $Y_c$ is pointed in the same direction as $Y_1$, and $X_c$ is defined along the imaginary cylinder central axis pointing towards the top of the imaginary cylinder, hence making an angle of $\phi$ with $X_1$ (See Fig. \ref{fig:Kinematics}(a)-(b)).\\
To determine any arbitrary point on the inner tube's neutral axis, the arc length parameter $s$ can take any values between 0 and $l_{na}$. Noting the directions of $O_c$ axes, the position vector expressed in $O_c$ coordinates ($\prescript{}{c}p(s)$) is obtained as a function of arc length $s$.
\begin{equation} \label{p_c}
    \prescript{}{c}{\vec{p}(s)} = 
    \begin{pmatrix}
        \frac{sH}{l_{na}}  \\
        -R\cos(\frac{2\pi s}{l_{na}}) \\
        R\sin(\frac{2\pi s}{l_{na}})    
    \end{pmatrix}
\end{equation}
To express this position vector in coordinate frame $O_1$, one must apply a translation of $R$ towards $-Y_c$ and a rotation of $-\phi$ around $Y_c$.
\begin{equation} \label{p_1}
    \prescript{}{1}{\vec{p}}(s)=
    \begin{pmatrix}
        \cos(\phi) & 0 & -\sin(\phi) \\
         0 & 1 & 0\\
        \sin(\phi) & 0 & \cos(\phi)
    \end{pmatrix}
    (\prescript{}{c}{\vec{p}(s)} + R\hat{y})
\end{equation}
Note that the $\hat{y}$ represents the standard unit vector in $y$ direction ($\hat{y}=[0,1,0]^T$). Now, To obtain the vector expression in $O_0$ coordinates, only a rotation of $\theta$ around $X_1$ needs to be applied.
\begin{equation} \label{p_0}
    \prescript{}{0}{\vec{p}}(s)=
    \begin{pmatrix}
        1 & 0 & 0 \\
        0 & \cos(\theta) & -\sin(\theta) \\
        0 & \sin(\theta) & \cos(\theta)
    \end{pmatrix}
    \prescript{}{1}{\vec{p}(s)}
\end{equation}
The obtained vector $\prescript{}{0}{\vec{p}}(s)$ describes the forward kinematics of the ExoNav robot.
\subsection{FTL Motion} \label{subsec:FTL_kin}
\textcolor{black}{As explained in Section \ref{sec2:Design},} the roller gear's translational DoF provides the capability to move the inner tube with respect to the outer tube, hence changing the constrained proximal portion of the inner tube that remains unbent. We call this motion of the robotic tool ``progressive motion``. There is a subset of progressive motions that match FTL motion \textcolor{black}{characteristics}. To perform FTL motion with the proposed robot, roller gear rotates as the robotic tool is \textcolor{black}{translated} out of the outer tube in a way that the actuation angle $\theta$ stays constant. Simultaneously, tendon stroke $\Delta l$ is modulated to keep the deflection angle, $\phi$, constant. By locking the values of these two angles, the location of the imaginary cylinder remains unchanged during the progressive motion, and based on the natural behavior of helical curves, the body of the robotic tool will follow its tip, performing a helical FTL motion.\\
To include the progressive motion in the kinematic model, a progression factor $\eta$ is introduced, which determines the exposed portion of the robotic tool neutral axis length out of the outer tube $\prescript{}{p}{l_{na}}$ with respect of its whole length $l_{na}$, i.e. 
    $\prescript{}{p}{l_{na}} = \eta l_{na}$.
Note that $\prescript{}{p}{l_{na}}$, and consequently $\eta$, are directly controlled by the roller gear's translational actuator. As the \textcolor{black}{inner tube} is \textcolor{black}{translating} out of the outer tube, the roller gear input angle $\theta_{in}=2\pi \eta$ must change linearly with $\eta$ to prevent the actuation angle $\theta$ (Shown in Fig. \ref{fig:Kinematics}(b)) from changing, effectively keeping the \textcolor{black}{inner tube's} notches \textcolor{black}{exposed immediately} at the tip of the outer tube always facing the same direction as the inner tube moves out.
To maintain the constant deflection angle $\phi$, the progressive tendon length (the length of the tendon routed through the exposed portion of the robotic tool out of the outer tube) also needs to change linearly with the progression factor $\eta$:
$
    \prescript{}{p}{l_t}=\eta l_t
$
For FTL motion, the tip position is calculated following the same steps as Eqs. (\ref{p_c}), (\ref{p_1}), and (\ref{p_0}), simply by substituting $\eta$ as a stand-in for arc length $s$. This shows that the kinematic equations guarantee that in this type of progressive motion, the body of the ExoNav inner tube will follow the tip. So, by changing $\eta \in$ [$0$,$1$], the \textcolor{black}{inner tube} is expected to perform a FTL motion.
\vspace{-10pt}
\section{Experiments and Results} \label{sec4:Experiments}
\begin{figurehere}
\begin{center}
\centerline{\includegraphics[width=3in,keepaspectratio]{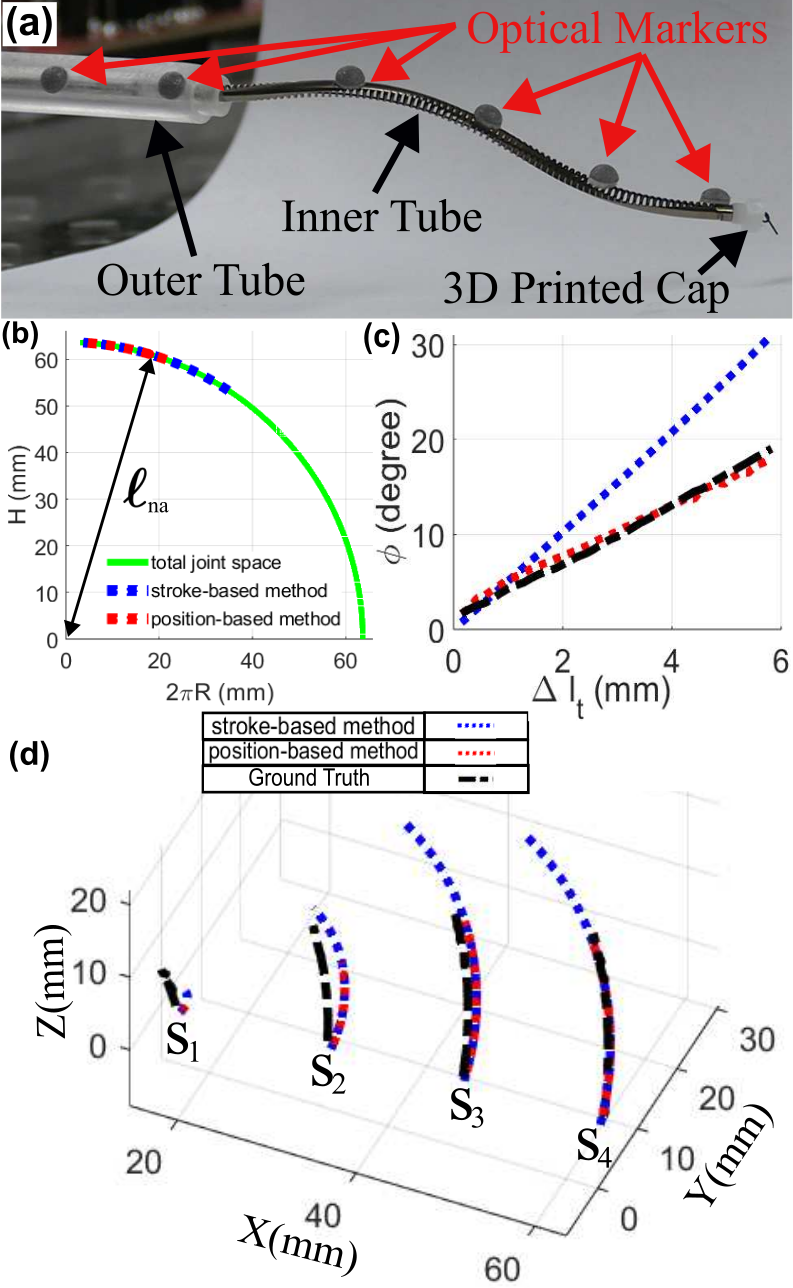}}
\caption{(a) Inner tube bending in a helical shape \textcolor{black}{with optical markers attached on it}, (b) Total joint space and joint space trajectory estimations, (c) Comparison of ground truth $\phi$ and model predictions, (d) Model-predicted marker trajectories compared to their actual measured positions.}
\label{fig:Exp1_Results}
\end{center}
\end{figurehere}
In this section, we describe two experiments to evaluate the \textcolor{black}{developed} kinematic model \textcolor{black}{in} Section \ref{sec3:kinematics}, compare the obtained experimental results with kinematic model's predictions, and report the corresponding accuracies. Ultimately, we present an \textcolor{black}{analysis of the robot's repeatability} and a demonstration of robot's operation in a 3D-printed spinal cord phantom model.
\subsection{Experiment 1: Inner Tube Shape}
\vspace{-10pt}
\begin{tablehere}
\tbl{Maximum Euclidean distance ($max(d_E)$) and RMSE values corresponding to markers position estimations by stroke-based and position-based methods. \label{Exp1_table}}
{\begin{tabular}{ |p{1.3cm}|p{1.5cm}|p{0.8cm}|p{1.5cm}|p{0.8cm}|  }
 \hline
  Marker & \multicolumn{2}{|c|}{stroke-based method} & \multicolumn{2}{|c|}{position-based method} \\
 \hline
 \centering $s$ (mm) & max($d_E$) (mm) & RMSE (mm) & max($d_E$) (mm) & RMSE (mm) \\
 \hline
 $18.24$ & $10.22$ & $8.17$ & $5.35$ & $4.43$ \\
 \hline
 $33.20$ & $19.84$ & $14.42$ & $10.54$ & $8.04$ \\
 \hline
 $48.05$ & $17.45$ & $8.70$ & $9.33$ & $5.40$ \\
 \hline
 $63.61$ & $14.27$ & $4.25$ & $7.52$ & $2.89$ \\
 \hline
\end{tabular}}
\end{tablehere}
\vspace{-10pt}
In the first set of experiments, the \textcolor{black}{inner tube} is completely translated out of the outer tube and is held at place without rotation, while the tendon is tensioned, bending the \textcolor{black}{inner tube} in a helical shape. We used Four Vero v2.2 motion tracking cameras (Vicon Motion Systems Ltd., United Kingdom) to track the four markers attached on the robot's body at approximate arc length values of $s_1 = 16.74$~mm, $s_2 = 33.20$~mm, $s_3 = 48.05$~mm, and $s_4 = 63.61$~mm \textcolor{black}{(where the helical notch pattern completes 25\%, 50\%, 75\%, and a complete rotation around the inner tube's circumference respectively)}. We attached four extra markers at fixed locations on the 3D-printed outer tube for \textcolor{black}{defining} the fixed global coordinate frame (See Fig. \ref{fig:Exp1_Results}(a)). After gathering all the experimental \textcolor{black}{results}, we obtained kinematic model estimations via two different methods: 1) stroke-based method and 2) position-based method.
In the \textbf{\textit{Stroke-based Method}}, we assume that only tendon stroke values are \textcolor{black}{available}. By substituting those values into Eq. (\ref{R-l_t relation}), we get $R$, and subsequently $H$ and $\phi$. However, in the \textbf{\textit{Position-based Method}}, we assume that the robot's end-effector position tracked by the marker at $s_4$ is known. Based on the kinematic model, the origin of the fixed global coordinate frame $O_0$ and the robot's end-effector will appear on the same side of the imaginary cylinder (See Fig. \ref{fig:Kinematics}(a)), implying that the angle between this vector and $+X_0$ is $\phi$ (taken as the ground truth), and the length of this vector is $H$. To \textcolor{black}{partially} evaluate the kinematic model, the obtain\textcolor{black}{ed} $H$ is substituted in Eqs. (\ref{l_na_R_H}) and (\ref{phi-R-H}) to obtain $R$ and $\phi$. \\
\textcolor{black}{Deflection angle} values obtained via both methods and the ground truth \textcolor{black}{were compared} (See Fig. \ref{fig:Exp1_Results}(c)) \textcolor{black}{which showed} close alignment between ground truth and position-based method \textcolor{black}{estimations, confirming} the assumption \textcolor{black}{of inner tube's helical shape upon actuation}. However, the $\phi$ obtained via stroke-based method is significantly off. This is likely due to the actuation dead-zones, and unknown elongation terms.\\
Using the parameters obtained from each method, we \textcolor{black}{estimated the} location of the markers along the \textcolor{black}{inner tube} using Eqs. (\ref{p_c}) and (\ref{p_1}). We suitably chose the actuation angle $\theta$ as a fixed value that confirmed the direction of robot's bending, and then it was substituted into Eq. (\ref{p_0}) to estimate the markers position vectors. We compared the \textcolor{black}{ground truth} with the stroke-based and position-based methods \textcolor{black}{estimations} (See Fig. \ref{fig:Exp1_Results}(d)). Furthermore, we calculated the maximum Euclidean distance \textcolor{black}{(denoted as $max(d_E)$)} between \textcolor{black}{these two sets of data for each method} and the \textcolor{black}{corresponding} RMSE values (See Table \ref{Exp1_table}). \textcolor{black}{Position-based method outperforms the stroke-based method, based on the reported error values.} The maximum RMSE value for estimations made by the position-based method is $8.04$~mm, which is 12.6\% of the robot's total length. \textcolor{black}{This RMSE value is still relatively high, possibly due to unmodeled tendon elongation, inner tube and outer tube frictions in angular direction, and overall actuators errors and backlashes.}
\begin{figure*}[t]
\centerline{\includegraphics[width=7in,keepaspectratio]{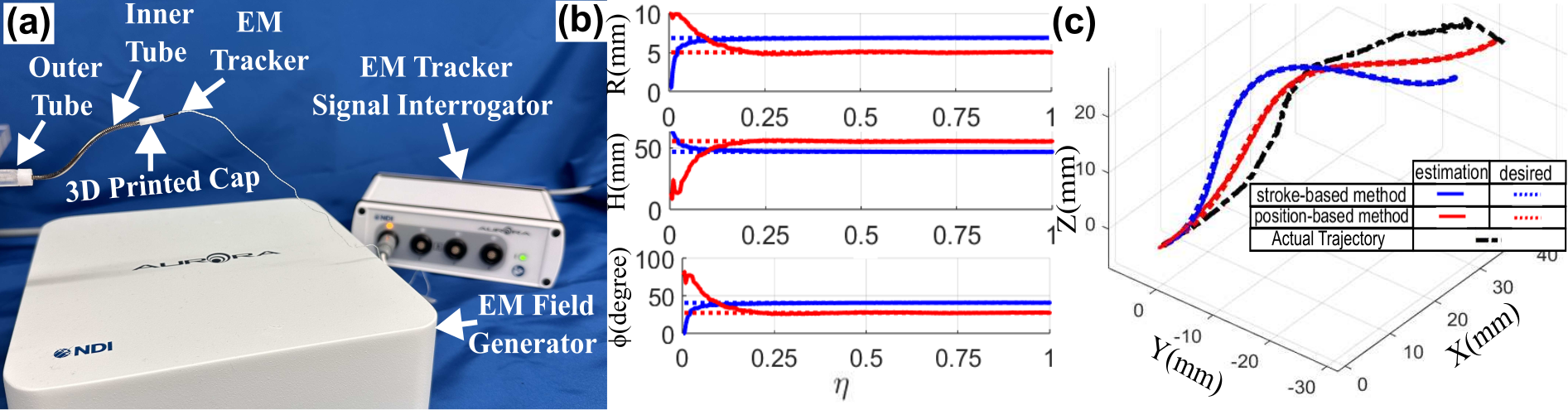}}
\caption{(a) Experimental setup for progressive motion experiments with electromagnetic tip-tracking, (b) Imaginary cylinder dimensions and deflection angles estimated by stroke-based and position-based methods showing steady constant values over high $\eta$, along desired FTL motion parameters. (c) Estimated, desired, and actual end-effector trajectories.}
\label{fig:Exp2_Results}
\end{figure*}
\subsection{Experiment 2: FTL Motion} \label{subsec:exp2}
\begin{tablehere}
\tbl{Maximum Euclidean distance ($max(d_E)$) and RMSE values corresponding to stroke-based and position-based methods estimating the robot's end-effector trajectory, and evaluation of the desired trajectories. \label{Exp2_table}}
{\begin{tabular}{ |p{1.5cm}|p{1.5cm}|p{0.8cm}|p{1.5cm}|p{0.8cm}|  }
 \hline
   & \multicolumn{2}{|c|}{stroke-based method} & \multicolumn{2}{|c|}{position-based method} \\
 \hline
  & max($d_E$) (mm) & RMSE (mm) & max($d_E$) (mm) & RMSE (mm) \\
 \hline
 Estimation & $11.45$ & $8.82$ & $7.08$ & $5.10$ \\
 \hline
 Desired & $11.24$ & $8.67$ & $7.32$ & $5.18$ \\
 \hline
\end{tabular}}
\end{tablehere}
\textcolor{black}{With the inner tube's helical shape aligning with forward kinematics assumptions and the helix's natural FTL motion capability, ExoNav is potentially capable of FTL motion. Therefore, experiment 2 is conducted to demonstrate this capability by comparing robot's end-effector trajectory with kinematic model predictions of a perfect FTL morion trajectory.} We utilized an electromagnetic (EM) tracking system (NDI Aurora, Northern Digital Inc, Waterloo, Ontario, Canada) \textcolor{black}{to measure the end-effector position in 3D space throughout this experiment. The EM tracker is attached to a 3D-printed cap component, redesigned for this purpose} (See Fig. \ref{fig:Exp2_Results}(a)). The progression factor $\eta$ is obtained based on the length of the exposed portion of the inner tube, measured by the translation actuator encoder. \textcolor{black}{Following the instructions presented in subsection \ref{subsec:FTL_kin}, input signals were generated to basically keep the imaginary cylinder at a fixed location and orientation throughout the experiment. The roller gear's translation actuator struggled to follow its reference signal during outward motion due to opposing tendon forces, but this issue was absent during retraction, so this experiment only shows FTL motion while the inner tube is retracted inside the outer tube. Using higher-power actuators or a more compliant inner tube (utilized later in subsection \ref{subsec:demo}) are possible solutions to the current outward progressive motion issue.}\\
We used the stroke-based and position-based methods to obtain imaginary cylinder dimensions and deflection angle values. As shown in Fig. \ref{fig:Exp2_Results}(b), both methods are showing steady constant values (labeled as desired) over a wide range of $\eta$, with some perturbations when $\eta$ gets close to zero (possibly due to the raised stiffness of the exposed portion of the inner tube and actuation tendon dead zones, which are expected to be relatively more significant for lower $\eta$ values). The desired \textcolor{black}{trajectories} describe a perfect FTL motion \textcolor{black}{(corresponding to the obtained joint space parameters)}. We \textcolor{black}{estimated} the position of the robot's end-effector \textcolor{black}{via stroke-based and position-based methods, and} used the desired joint space parameters to generate the desired \textcolor{black}{end-effector} trajectory. \textcolor{black}{These} trajectories were then compared to the actual \textcolor{black}{end-effector} position. \textcolor{black}{Note that comparison between the actual and ``estimated`` trajectories illustrate kinematic model evaluation in progressive motion and does not include evaluation of adherence to a FTL motion. However, alignment of actual and either one of ``desired`` trajectories show the robot's FTL motion capability and the kinematic model ability to characterize that FTL motion (See Fig. \ref{fig:Exp2_Results}(c)).} We reported $max(d_E)$ and RMSE values for \textcolor{black}{both methods} in Table \ref{Exp2_table}. As expected, the position-based method estimations \textcolor{black}{show} better alignment with actual data compared to the stroke-based model (See the first row in Table \ref{Exp2_table}). \textcolor{black}{Similarly, desired FTL trajectory, obtained by position-based method, lies closer to the actual trajectory of the robot's end-effector. However, due to open-loop operation of the robot, significant errors are reported. Future studies will focus on compensating these errors by utilizing feedback control systems.}
\textcolor{black}{ \subsection{Repeatability Analysis}
\begin{figurehere}
\begin{center}
\centerline{\includegraphics[width=3in,keepaspectratio]{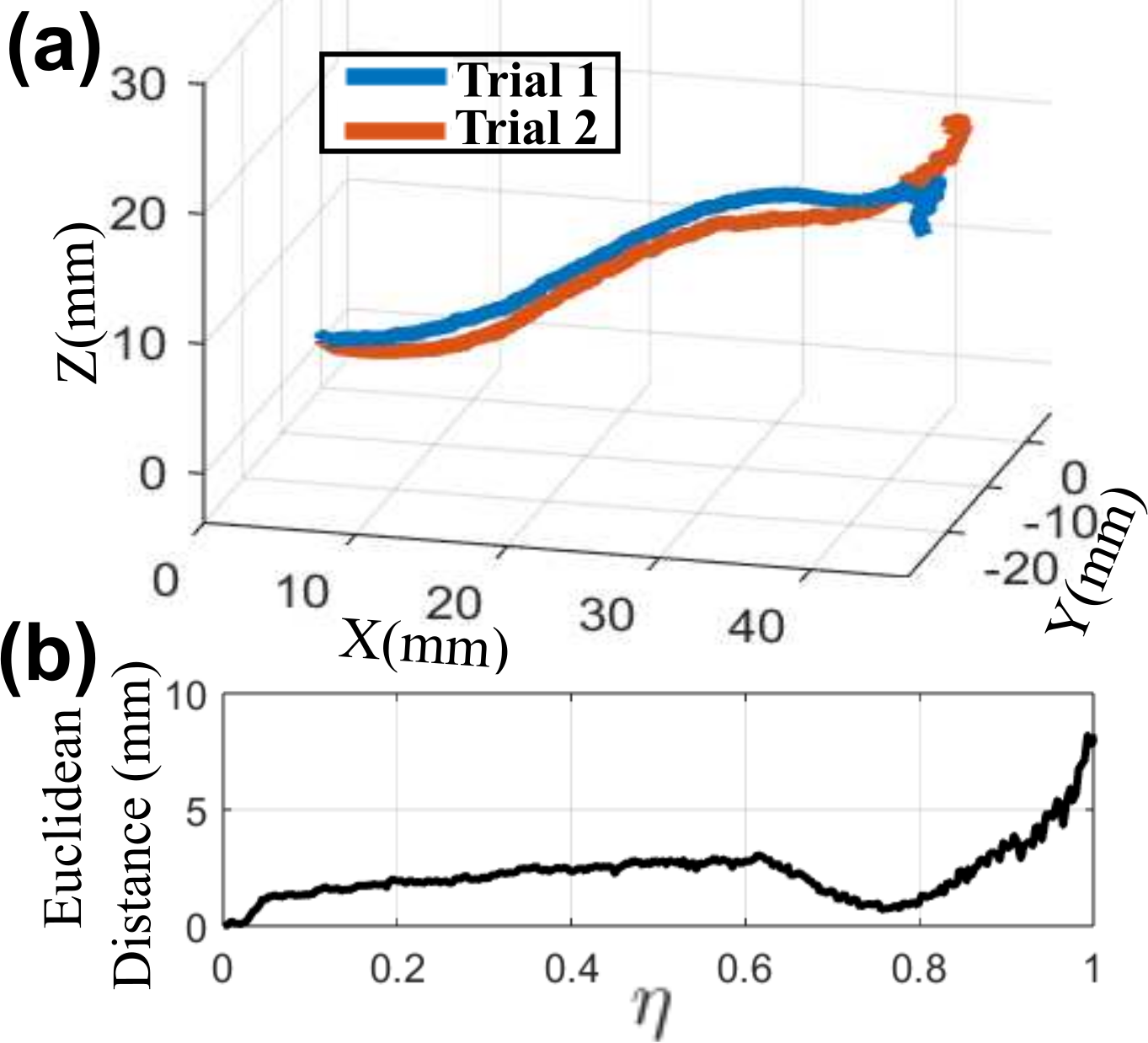}}
\caption{(a) Robot's end-effector trajectories in two different FTL motion experimental trials. (b) The Euclidean distance error plotted over the full range of progression factor values.}
\label{fig:Repeatability}
\end{center}
\end{figurehere}
\vspace{-15pt}
To illustrate the repeatability of the presented results in previous subsections, we demonstrate the close alignment of end-effector trajectories from two different trials of the FTL motion experiment (See Fig. \ref{fig:Repeatability}(a)). The maximum Euclidean distance is $8.23$~mm, which occurs at $\eta = 1$, most probably due to unmodeled friction sources between inner and outer tubes. However, on a wide range of smaller $\eta$ values, the error remains significantly low (See Fig. \ref{fig:Repeatability}(b)), which is shown by RMSE value of $2.62$~mm.}
\textcolor{black}{\subsection{Demonstration in Phantom Model} \label{subsec:demo}
In} our final step, we 3D-printed a spinal cord phantom out of Ninjaflex (Ninjatek) thermoplastic polyurethane (TPU) filament on a Bamboo lab X1-Carbon 3D-printer. We placed the spinal cord phantom in a suitable angle and direction in ExoNav outer tube\textcolor{black}{'s tip} proximity. Additionally, we used a digital camera (Dino-Lite digital microscope, Dunwell Tech, Inc.) to record videos (See Fig. \ref{fig:Demo}(a)). \textcolor{black}{To resolve the issue with outward progressive motion (stated in subsection \ref{subsec:exp2}), we utilized a smaller and more compliant prototype of the inner tube (with specifications of $R_{out}=0.57$~mm, $R_{in}=0.46$~mm, $H=2.15$~mm, $w=0.4$~mm, $R_d=0.2$~mm, and $l=75.4$~mm).} Based on visual assessment \textcolor{black}{of the inner tube's body trajectory, the inner tube successfully conducted a FTL motion} with minimal interference with the spinal cord phantom (See Fig. \ref{fig:Demo}(b)).
\begin{figurehere}
\begin{center}
\centerline{\includegraphics[width=3in,keepaspectratio]{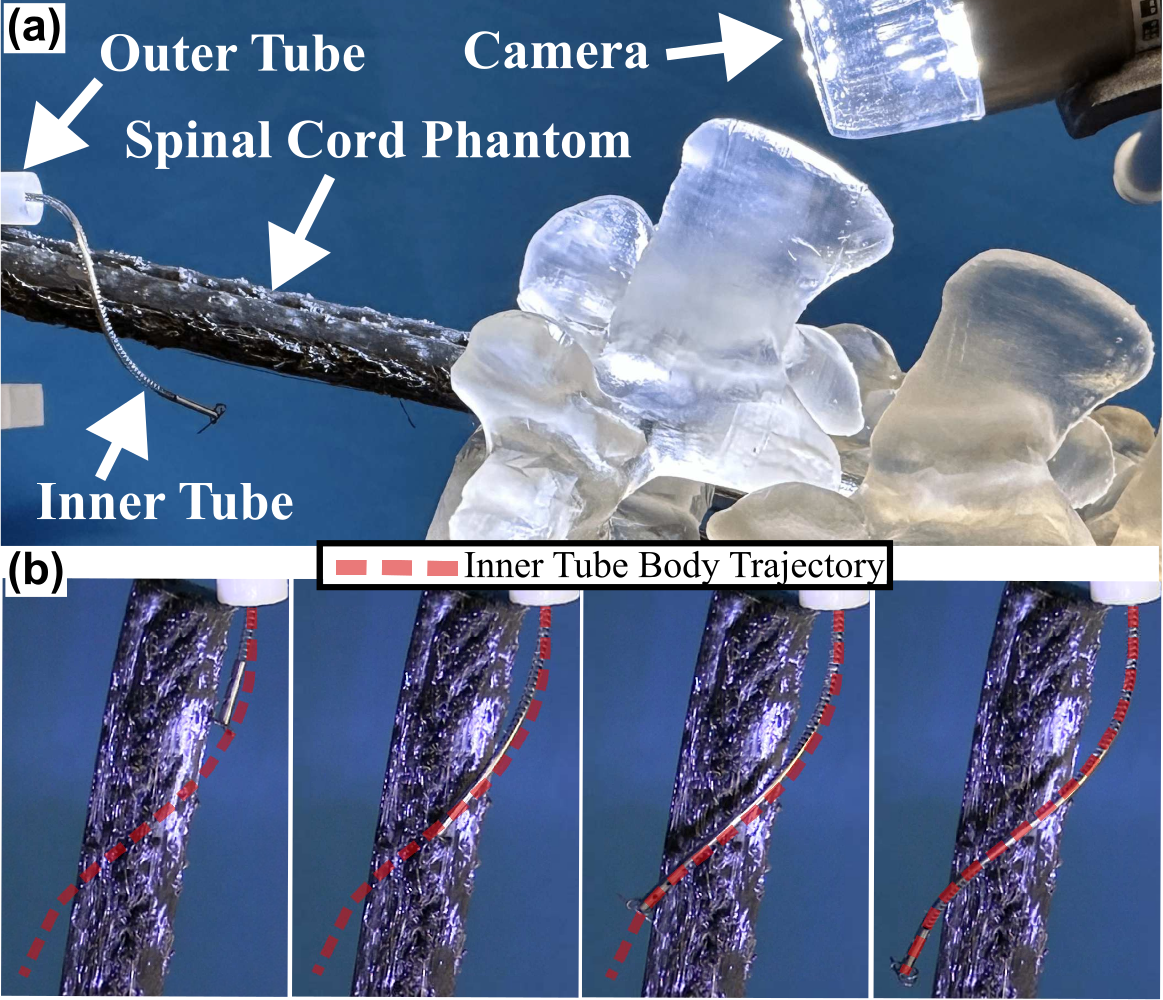}}
\caption{(a) Spinal cord phantom demonstration setup. (b) the evolution of the inner tube FTL motion performed around the 3D-printed spinal cord phantom (from left to right) \textcolor{black}{with end-effector closely following the final body trajectory.}}
\label{fig:Demo}
\end{center}
\end{figurehere}
\vspace{-35pt}
\section{Conclusions and Future Work} \label{sec5:conclusion}
This manuscript introduces a novel continuum robot design \textcolor{black}{directly targeting the ventral and lateral motor tracts in the spinal cord, intended to improve motor recovery after spinal cord injury}. The robot’s design and kinematics, based on geometric principles and conventional modeling paradigms, enable it to perform a repeatable helical FTL motion, which was confirmed experimentally and demonstrated in a spinal cord phantom model. \textcolor{black}{Future work will enhance the ExoNav robot's accuracy and robustness (by optimizing navigation with improved modeling and feedback systems) and explore integrating SCS electrodes or latching mechanisms.}
\nonumsection{Acknowledgments}
\noindent This work was supported in part by the Kentucky Spinal Cord and Head Injury Research Trust (KSCHIRT) Fund.
\bibliographystyle{IEEEtran}
\bibliography{references}

\noindent\includegraphics[width=1in]{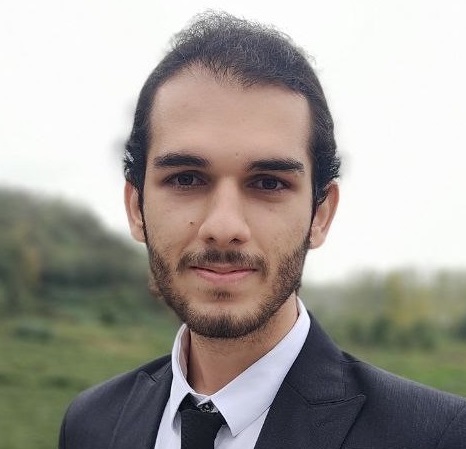}
{\bf Behnam Moradkhani} received the B.Sc. degree in electrical engineering from the University of Tehran, Tehran, Iran, in 2022. He is currently working toward the Ph.D. degree in mechanical engineering with the University of Louisville, Louisville, KY, USA. His research focuses on robotics and control systems.\\

\noindent\includegraphics[width=1in]{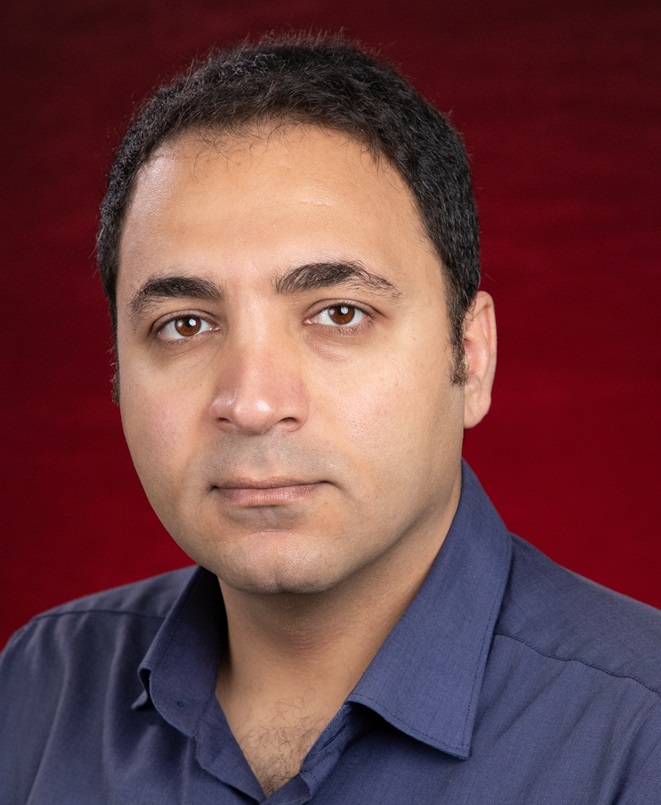}
{\bf Pejman Kheradmand} received the bachelor’s degree in mechanical engineering from Razi University, Kermanshah, Iran, and the master’s degree in mechanical engineering from the University of Tehran, Tehran, Iran. He is currently working toward the Ph.D. degree in mechanical engineering with the University of Louisville, Louisville, KY, USA. His research focuses on continuum robots in surgical contexts.\\

\noindent\includegraphics[width=1in]{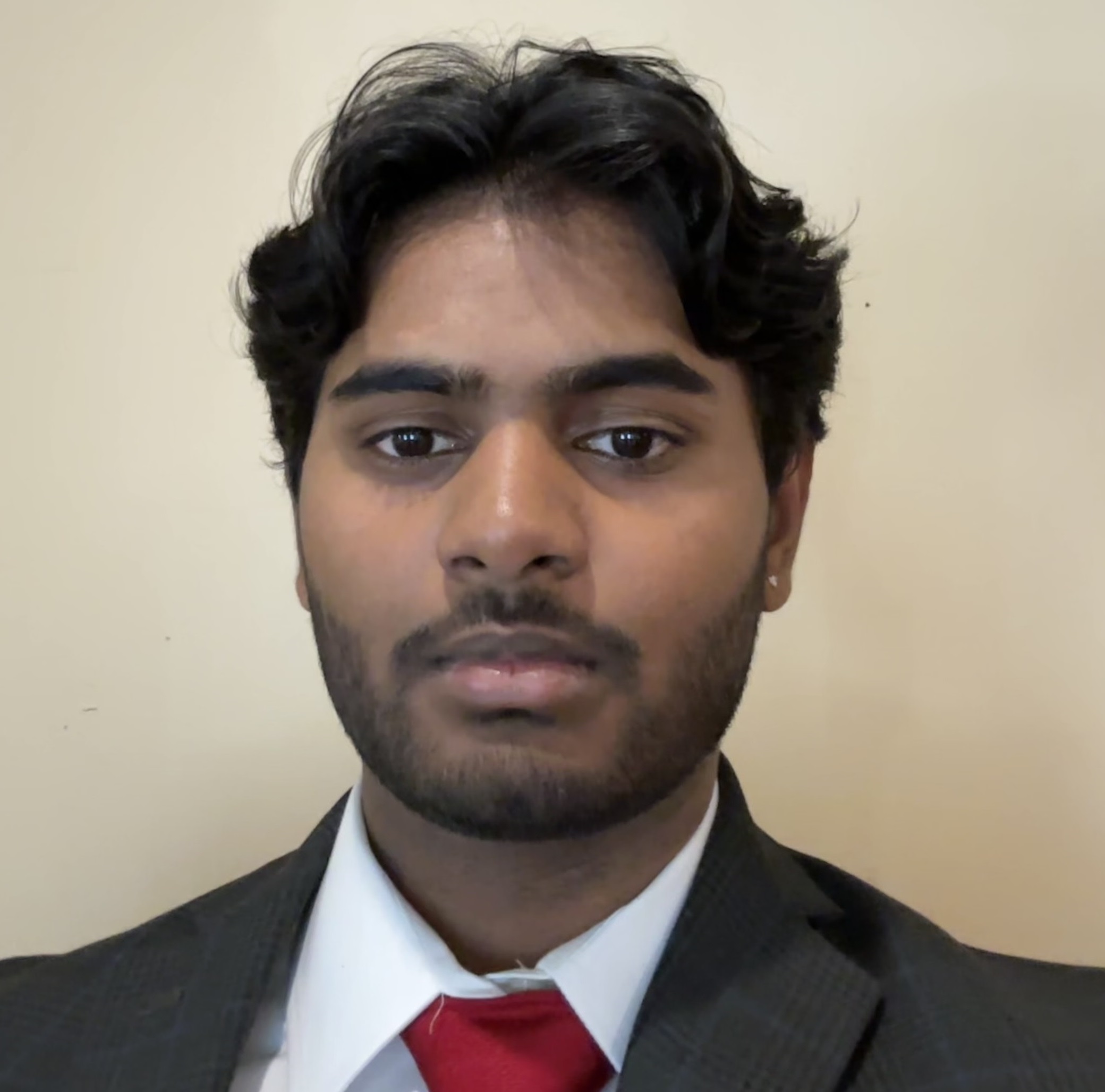}
{\bf Harshith Jella} is currently working toward the B.S. degree in biology with the University of Louisville, Louisville, KY, USA. His research interests include medical robotics and oncolytic virotherapy.\\

\noindent\includegraphics[width=1in]{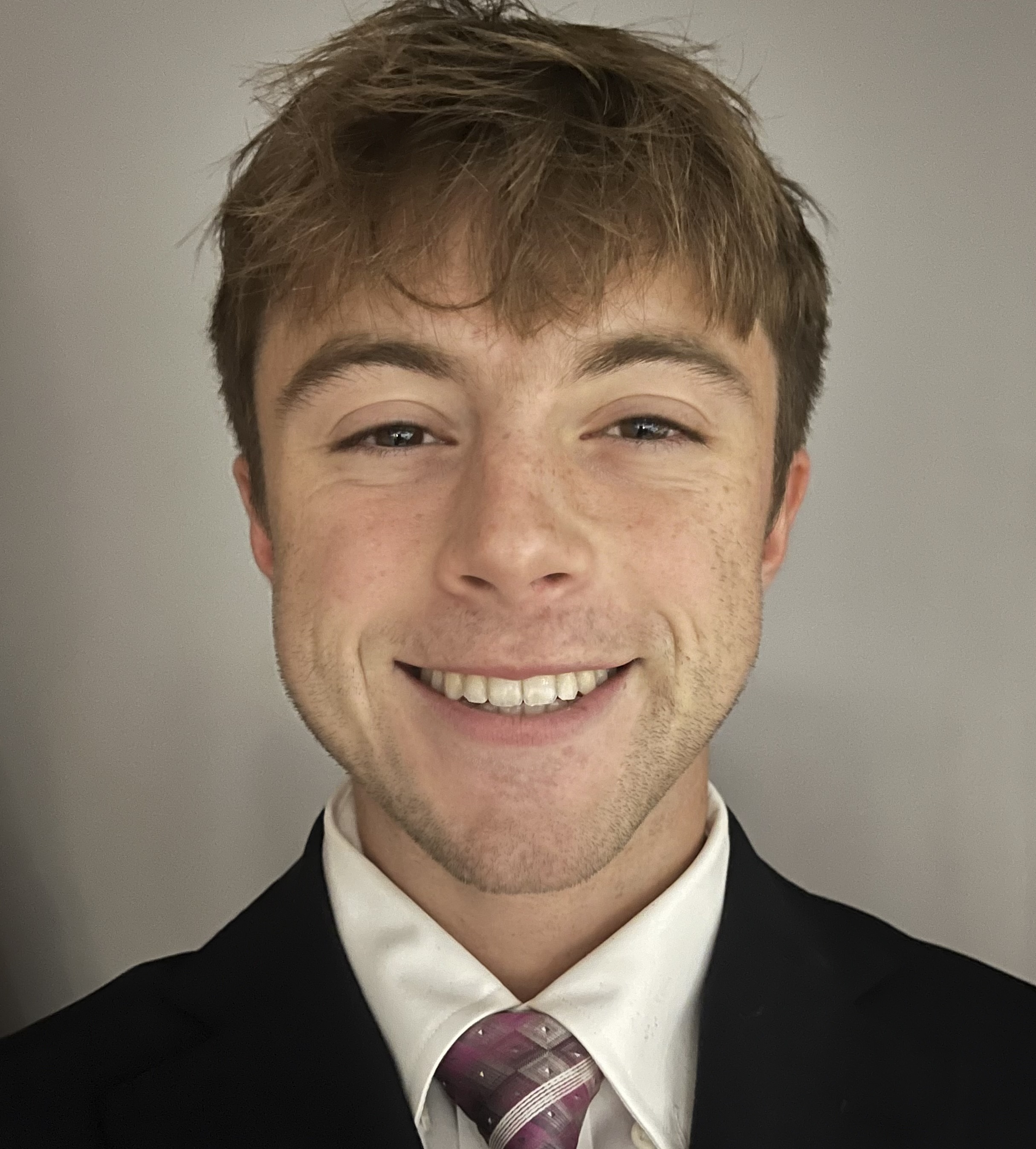}
{\bf Joseph Klein} is currently working towards a B.S. degree in mechanical engineering with a minor in computer science from the University of Louisville, Louisville, KY, USA. His research mainly focuses on biomechanics and medical robotics.\\

\noindent\includegraphics[width=1in]{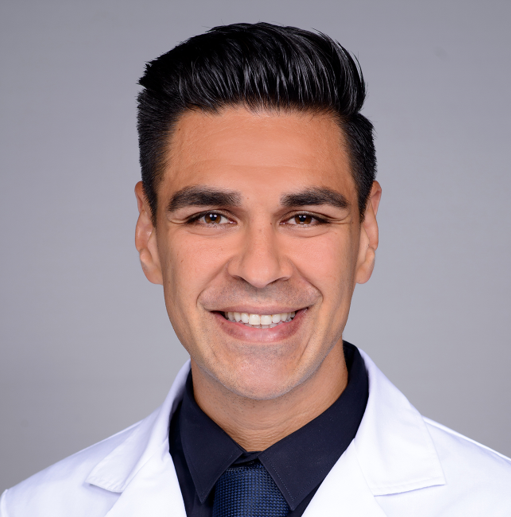}
{\bf Ajmal Zemmar} received the medical degree in neurosurgery from Goethe University, Frankfurt, Germany, and the Ph.D. degree in neuroscience from the Swiss Federal Institute of Technology, Zurich, Switzerland. He is a dual-trained Physician–Scientist Neurosurgeon. He completed neurosurgery residency in Zurich (Switzerland) and Vancouver (Canada), along with a fellowship in functional neurosurgery in Toronto (Canada). His research focuses on developing magnetic robotic technologies for curved trajectories in neurosurgery. He has authored or coauthored more than 60 peer-reviewed articles, five book chapters, multiple academic grants, awards, and patents, Dr. Zemmar is an Editor and a Reviewer for international journals, the European Research Council, and the Michael J. Fox Foundation. He is the Chief Medical Officer of the National Parkinson’s Institute and a Member of the Congress of Neurological Surgeons, American Association of Neurological Surgeons, European Association of Neurosurgical Societies, and the American Medical Association.\\

\noindent\includegraphics[width=1in]{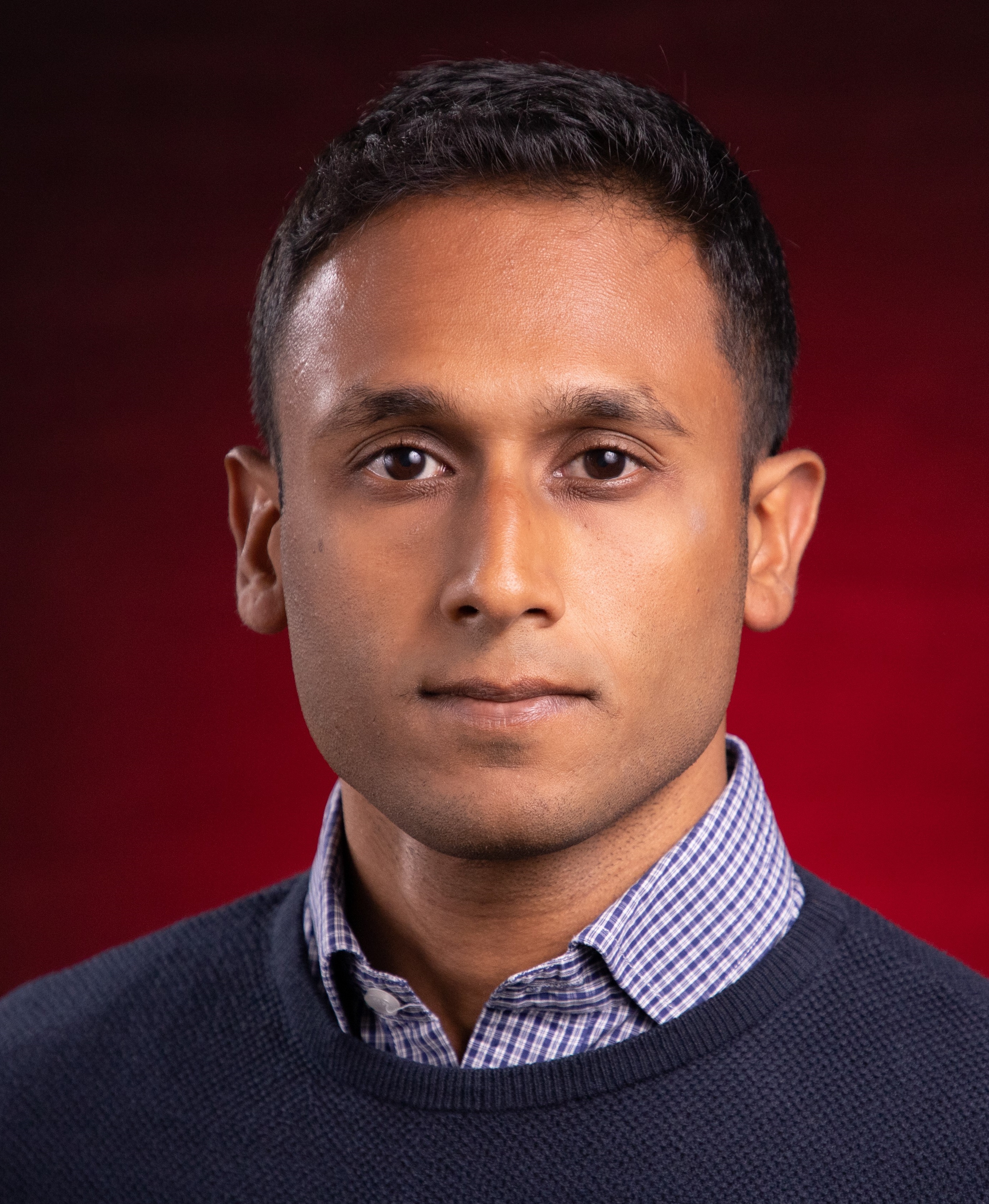}
{\bf Yash Chitalia} (Member, IEEE) received the bachelor’s degree in electronics engineering from the University of Mumbai, Mumbai, India, in 2011, the M.S. degree in electrical engineering from the University of Michigan, Ann Arbor, MI, USA, in 2013, and the Ph.D. degree in mechanical engineering from the Georgia Institute of Technology, Atlanta, GA, USA, in 2021. He is currently an Assistant Professor with the Department of Mechanical Engineering, University of Louisville, Louisville, KY, USA. His research interests include microscale and mesoscale surgical robots for cardiovascular and neurosurgical applications as well as rehabilitation robotics. Dr. Chitalia was the recipient of several awards including the Ralph E. Powe Junior Faculty Enhancement Award and the NASA-KY Research Initiation Award.

\end{multicols}
\end{document}